\documentclass{article}

\usepackage[english]{babel}

\usepackage[letterpaper,top=2cm,bottom=2cm,left=3cm,right=3cm,marginparwidth=1.75cm]{geometry}
\usepackage{amsmath}
\usepackage{amssymb}
\usepackage{CJKutf8}
\usepackage{graphicx}
\usepackage{appendix}
\usepackage{adjustbox}
\usepackage{tabularx}
\usepackage{soul} 
\usepackage{color, xcolor} 
 
\usepackage[
    style=numeric-comp,
    sorting=none
]{biblatex}
\title{\textbf{PatentGPT: A Large Language Model for Intellectual Property}
}
\author{Zilong Bai, Ruiji Zhang, Linqing Chen, Qijun Cai, Yuan Zhong, Cong Wang \\
Yan Fang, Jie Fang, Jing Sun, Weikuan Wang, Lizhi Zhou, Haoran Hua\\ 
Tian Qiu, Chaochao Wang, Cheng Sun, Jianping Lu, Yixin Wang, Yubin Xia \\
Meng Hu, Haowen Liu, Peng Xu, Licong Xu, Fu Bian, Xiaolong Gu, Lisha Zhang \\
Weilei Wang\footnote{Corresponding author}, Changyang Tu\\
Patsnap LLM Team\\
baizilong@patsnap.com, wangweilei@patsnap.com
}
\definecolor{lightblue}{rgb}{0.8,0.8,0.8}
\begin{document}
\maketitle
\setlength{\parindent}{0pt}
\begin{abstract}
\noindent
In recent years, large language models(LLMs) have attracted significant attention due to their exceptional performance across a multitude of natural language process tasks, and have been widely applied in various fields. However, the application of large language models in the Intellectual Property (IP) domain is challenging due to the strong need for specialized knowledge, privacy protection, processing of extremely long text in this field. In this technical report, we present for the first time a low-cost, standardized procedure for training IP-oriented LLMs, meeting the unique requirements of the IP domain. Using this standard process, we have trained the PatentGPT series models based on open-source pretrained models. By evaluating them on the open-source IP-oriented benchmark MOZIP, our domain-specific LLMs outperforms GPT-4, indicating the effectiveness of the proposed training procedure and the expertise of the PatentGPT models in the IP domain. Remarkably, our model surpassed GPT-4 on the 2019 China Patent Agent Qualification Examination, scoring 65 and matching human expert levels. Additionally, the PatentGPT model, which utilizes the SMoE architecture, achieves performance comparable to that of GPT-4 in the IP domain and demonstrates a better cost-performance ratio on long-text tasks, potentially serving as an alternative to GPT-4 within the IP domain.

\end{abstract}

\section{Introduction}
Over the past few years, Large Language Models (LLMs) based on decoder-only transformers have received substantial attention, due to their excellent alignment with human preference and remarkable performance across various NLP tasks, including reading comprehension, novel creation, text summarization, code completion, and document drafting. Recent research suggests that LLMs are not merely reproducing surface statistics, but are instead learning a meaningful world model. \cite{1,2,3} opening up the imaginative space for the application of LLMs. The performance of LLMs follows the scaling law that models with more pretraining data and parameters achieve better performance \cite{4,5,6,7}, indicating that the costs to train and apply a high-performance LLM are significant.\\
\\
To reduce the pretraining and deployment costs of LLMs, more and more pretrained models are being published in open-source communities. These LLMs, including LLaMA \cite{8,9}, Bloom \cite{10}, and ChatGLM \cite{11}, have parameters ranging from several billions to tens of billions, and they were pretrained on over a trillion tokens, showing exceptional performance in base capabilities. Developers can adapt these general models into domain-specific models by continuous training or fine-tuning without conducting an extensive pretraining process. We have selected LLama as the base model for further pretraining among open-source LLMs. These models, which range from 7 billion to 70 billion parameters in LLama2 and from 8 billion to 70 billion parameters in LLama3, integrate various significant works to establish an effective and stable framework. Consequently, the LLaMA series has established itself as the standard reference for architecture and performance within open-source models.\\ 
\\
Low-resource inference for LLMs also attracts more attention. Main research focuses on implementation of quantization \cite{12,13,14,15,16,17}, knowledge distillation \cite{18,19,20,21,22}, and sparse activation mechanisms \cite{23,24,25,26}. Recently, Jiang et al. research \cite{27} on Sparse Mixture-of-Experts (SMoE) has demonstrated that compared to dense LLMs of the same parameter scale, SMoE LLMs allow faster inference speed and higher throughput while maintaining comparable performance.Their findings indicate that SMoE is one of the most promising architectures for reducing inference resource requirements. By sharing attention heads between experts and reducing the dimensions of key and value tensors within the attention heads, SMoE LLMs can significantly reduce the memory consumption of the key-value Cache (KV cache) when processing long texts, thereby providing better support for long-document applications. This is particularly important in long-text comparison scenarios.\\ 
\\
Furthermore, it is worth exploring how cost-effective and high-performance LLMs can be implemented within domain-specific industries. Recent research \cite{28,29,30,31} suggests that by continuous training of generic LLMs on domain-specific data, smaller models can achieve the same or even better performance in specialized fields compared to much larger models with significantly more parameters. This result provides a new approach for cost-effective and specialized applications of LLMs in specific domains. For the IP domain, the main applications of LLMs are concerned with responding to IP-oriented legal questions, assisting in reading patent specifications, drafting patent documents, and comparing patent documents in the context of Freedom-To-Operate (FTO) scenarios. These applications either require specialized knowledge, introduce privacy concerns, or involve extremely long input texts that lead to high practical costs. Therefore, it is unwise to directly apply open-source models (e.g., LLaMA) or commercial models (e.g., GPT-4) to the IP domain.\\
\\
In this report, we introduce methodologies for training IP-oriented LLMs and establish a benchmark called PatentBench in the IP domain. By training on over 240 billion tokens of IP-oriented data, our PatentGPT models exhibit performance in the IP domain that surpasses that of GPT-4. Further experiments suggest that the PatentGPT model with SMoE architecture shows a better cost-performance ratio on long-text tasks compared to dense LLMs.

\section{Pretraining}

We trained PatentGPT models with the base models of LLaMA2 (13B), LLaMA2 (70B) \cite{9}, and Mixtral $8\times7B$ \cite{27}, and we named the series of LLMs PatentGPT-0.5, PatentGPT-1.0-Dense and PatentGPT-1.0-MoE respectively.

\subsection{Pretraining Data}

\begin{figure}[!h]
\centering
\includegraphics[width=0.6\textwidth]{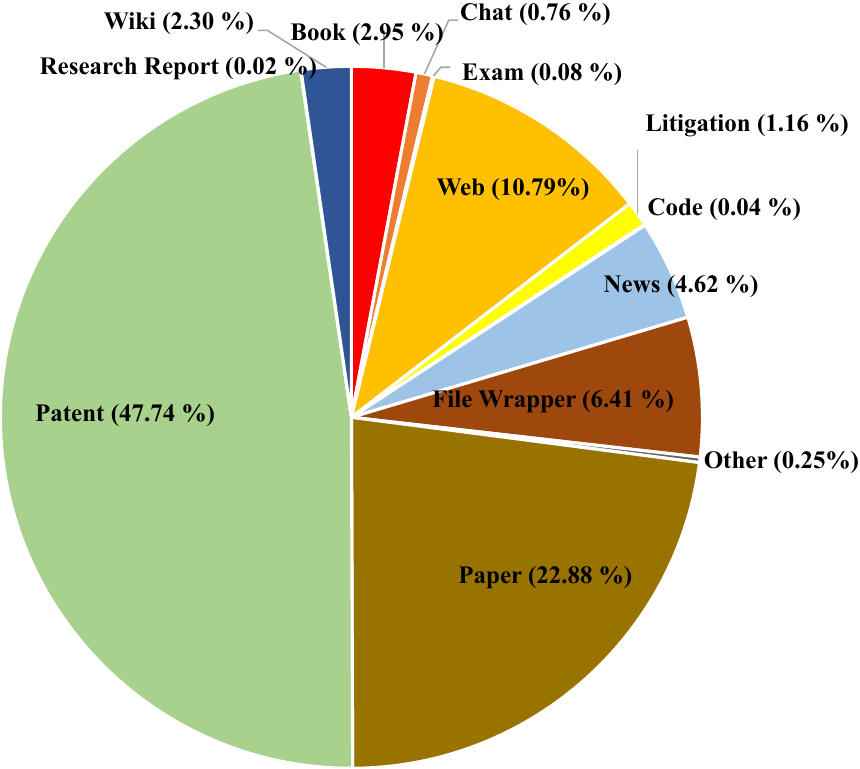}
\caption{\label{fig1} The distribution of different categories of pretraining data for PatentGPT models.}
\end{figure}

To create an effective pre-training dataset, it is crucial to ensure that the pretraining data are diverse and cover a wide range of types, domains, and tasks. In the IP domain, it involves legal knowledge as well as engineering and technical knowledge from various disciplines, and requires a strong ability to contextualize when writing patent specifications and comparing distinctions between two patents. Aiming to meet these requirements, the pretraining dataset was made up of publicly available internet resources, such as websites, wikis, books, exam databases, code repositories, and news articles, as well as non-public internal sources, including patents, file wrappers \cite{32}, annotated datasets, and litigation records. Additionally, we incorporated data from third parties like research papers, books, and research reports, along with internally generated data. We trained PatentGPT models using a two-stage pretraining process over 240 billion tokens of data in both English and Chinese. The proportion of Chinese pretraining data is 37\%. The composition of the training data is shown in Figure \ref{fig1}.
To obtain a high-quality dataset for pretraining, we employed several data preprocessing procedures including filtering, deduplication rewriting, and synthesis.\\ 
\\
\textbf{Filtering:} A hybrid approach of rule-based and machine-learning-based techniques was engaged to remove toxicity and low-quality data, including data feature filtering logic set by business experts, binary classification-specific mini-models constructed by our technical teams, black/white list keyword filters, and other internal data cleansing tools.\\ 
\textbf{Deduplication:} Both exact matches and approximate matches were used to remove repeated documents from the dataset. For exact matches, we matched substrings within the documents and their corresponding attribute labels, including the Digital Object Identifier (DOI) for academic papers, as well as the patent family information for patents. For fuzzy duplicate detection, MinHash \cite{33} was utilized during data preprocessing.\\ 
\\
\textbf{Rewriting:} Inspired by Li et al. \cite{34,35}, we up-sampled data from books with IP-oriented knowledge about terminologies and regulations of IP systems by rewriting the original documents into various expressions, including summarization and conversation, to inject IP-oriented knowledge into LLMs more efficiently.\\ 
\\
\textbf{Synthesis:} In practice, most LLMs struggle in differentiating patents or retrieving similar patents given the large amount of candidates and high overlapping of text. To enable PatentGPT models to better understand the differences and inherent connections between similar patents, we have reorganized the data from different sources through their attribute labels. A file wrapper is a written record in a patent office of the application and negotiations for a patent preceding the issuance of the patent. To obtain synthesis data, we concatenated the claims of the patent under examination with the full text of the corresponding comparative patents. We then added a file wrapper document at the end to reveal the connections between the two patents. We also recombined a document by concatenating two patents that share an X-File relationship to facilitate the model in understanding the connections between the two patents. Data examples can be found in Appendix \ref{A}.
\subsection{Training Details}
We began with the pretraining approach described in Touvron et al. \cite{9}, using an optimized auto-regressive transformer. However, we made several changes to pretrain IP-oriented LLMs. These changes include multilingual pretraining and a carefully constructed, IP-oriented pretrain dataset as described in Section 2.1. Additionally, we conducted a two-stage pretraining process to inject IP-oriented knowledge as well as develop abilities for recognizing, drafting, and comparing patents. All PatentGPT models were trained on GPU clusters using Megatron-LM \cite{36} equipped by flash attention modules. To create batches of data, we shuffled and merged the documents, and then concatenated or truncated them to the specified context lengths. We adopt the standard optimizer AdamW \cite{37} for pretraining optimization and trained the LLMs with BFloat16 mixed precision for training stability. We used a cosine learning rate schedule, and decayed final learning rate down to 10\% of the peak learning rate. The number of training tokens in each batch were set to 4 M. The attributes and the corresponding two-stage pretraining hyperparameters of PatentGPT models are listed in Table \ref{table1}.\\
\begin{table}[!h]
\footnotesize
\centering
\begin{tabular}{p{2.9cm}|p{1.8cm}p{0.9cm}p{0.9cm}p{0.9cm}p{1.1cm}p{1.1cm}|p{1.1cm}p{1.1cm}}
\hline
\multicolumn{1}{c}{} & \multicolumn{4}{|c}{} & \multicolumn{2}{c|}{\textbf{Stage 1}} & \multicolumn{2}{c}{\textbf{Stage 2}} \\\rule{0pt}{2ex}
 & \textbf{Base Model}&\textbf{Params}&\textbf{Tokens}&\textbf{LR}&\textbf{Context Length}&\textbf{Warmup}&\textbf{Context Length}&\textbf{Warmup} \\
\hline
\rule{0pt}{3ex}PatentGPT-0.5 & LLaMA 2 & 13B & 246B& 2$\times10^{-5}$ & 4K & 2K& 16K & 1K \\
PatentGPT-1.0-Dense & LLaMA 2 & 70B & 246B& 2$\times10^{-5}$ & 4K & 2K& 4K & 1K \\
PatentGPT-1.0-MoE & Mixtral 8$\times$7B & 47B & 246B& 3$\times10^{-5}$ & 8K & 2K& 16K & 1K \\
\hline
\end{tabular}
\caption{\label{table1}The attributes and the corresponding two-stage pretraining hyperparameters of PatentGPT models.}
\end{table}

\begin{figure}[!h]
\centering
\includegraphics[width=0.8\textwidth]{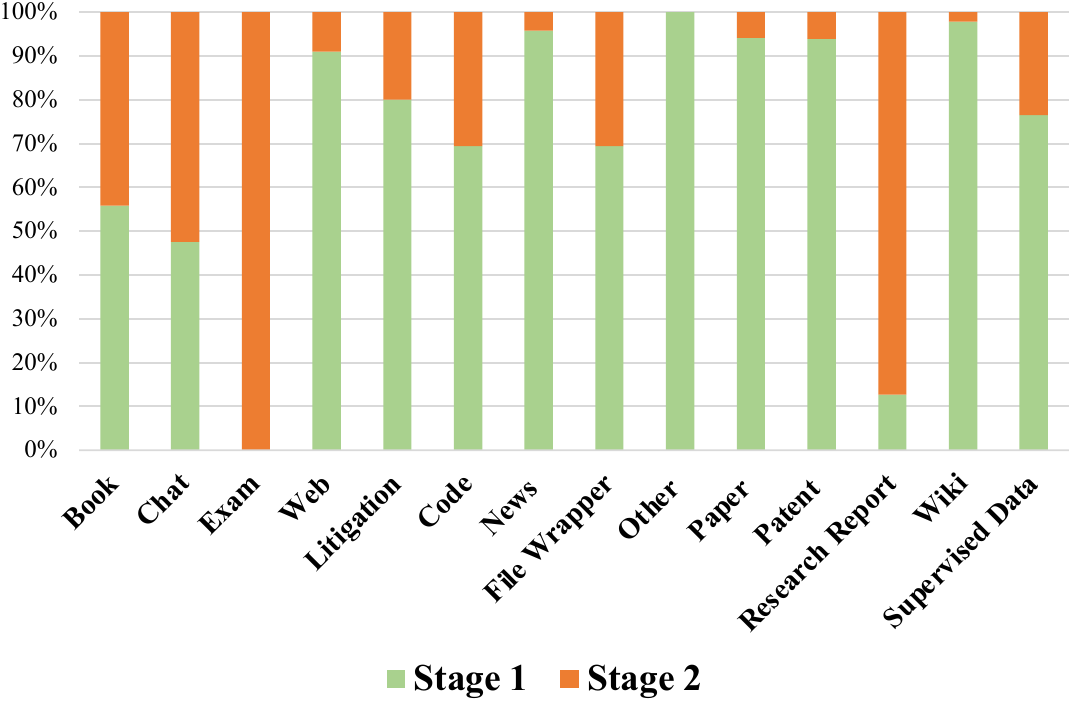}
\caption{\label{fig2} The proportion of different types of data used in each pretraining stage compared to the total amount of the corresponding type of pretraining data.}
\end{figure}

\textbf{Multilingual pretraining:} Before the pretraining process, we trained a new tokenizer using byte-pair encoding (BPE) \cite{38} from SentencePiece \cite{39} on the pretraining data. Then, we merged it with the LLaMA2 tokenizer to obtain a new tokenizer with a vocabulary size of 55,296, aiming to improve the token compression rate for text in Chinese and the IP domain. The new tokenizer contains an additional 23,296 tokens to the LLaMA2 tokenizer and is shared by all PatentGPT models. To adapt our PatentGPT models for the new tokenizer, we resized the word embedding layers and output layers from shape $V\times{H}$ to $V’\times{H}$, where V = 32,000 denotes the original vocabulary size, and V = 55,296 is the new vocabulary size of the new tokenizer. The new rows were appended to the end of the original embedding matrices, ensuring that the embeddings of the tokens in the original vocabulary remain unaffected. Finally, the additional parameters were trained in the further pretraining process. This multilingual alignment process ensures the enhancement of the context compression ratio, Chinese contextual understanding, and generation capabilities of our PatentGPT models.\\ 
\\
\textbf{Two-stage pretraining:} We trained the PatentGPT models using a two-stage pretraining process, during which 226B tokens were consumed in stage 1 and 20B tokens in stage 2, respectively. Figure \ref{fig2} shows the proportion of different types of data used in each pretraining stage compared to the total amount of the corresponding type of pretraining data. The data of different categories is not randomly distributed in the pretraining processes of the two stages. During stage 1, almost all the data from Web, News, Patent and Paper were consumed to inject basic knowledge into the PatentGPT models. For stage 2, most of the data from Research Reports and Exams, and a considerable proportion of the data from Books, Chats, Codes, File Wrappers and Supervised Data were used. It should be noted that due to the total tokens of pretraining data in stage 2 being less than a tenth of that in the stage 1, the proportion of data from Books, Chats, Exams, Codes, File Wrappers, and Research Reports relatively increased in stage 2, ensuring our PatentGPT models can be adequately trained in IP-oriented knowledges and tasks.\\
\\
\textbf{Long context extending:} As expected to reduce computation resources, we trained PatentGPT-0.5 and PatentGPT-1.0-MoE in a shorter context length in stage 1, and then extended them to 16k during stage 2. For PatentGPT-0.5, the context length of the base model is 4k. Inspired by Xiong et al. \cite{40}, we modified the base frequency of PatentGPT-0.5 from 10k to 100k and extended the training context length from 4k to 16k in stage 2, aiming to reduce the model performance regression. Due to the fact that the original context length of Mixtral $8\times7B$ is 32k, we directly concatenated the samples in pretraining data to 16k and trained the model in stage 2, without performing any other special conductions.

\section{Alignment}
\begin{figure}[!h]
\centering
\includegraphics[width=0.8\textwidth]{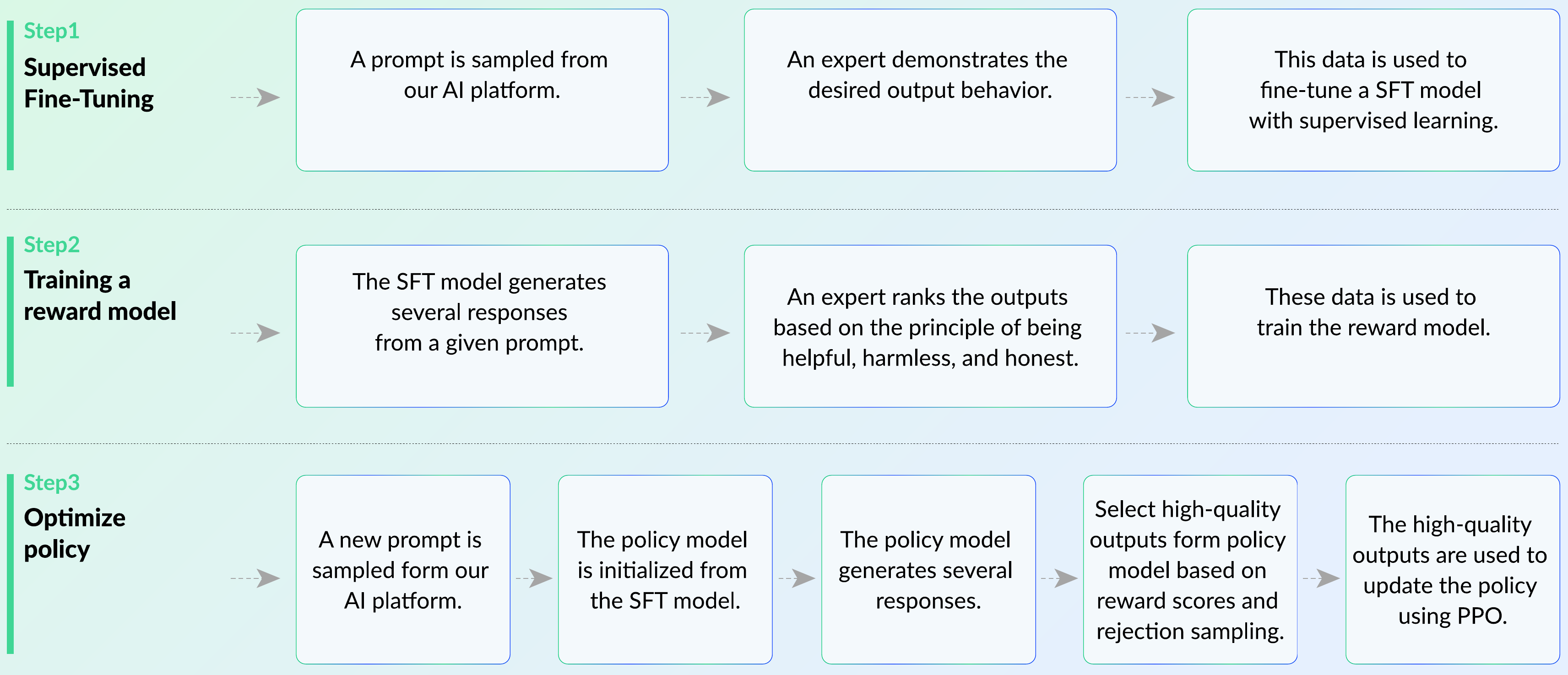}
\caption{\label{fig3} The workflow for conducting SFT and RLHF on PatentGPT models}
\end{figure}

LLMs have demonstrated impressive performance in various NLP tasks. However, their implementation presents limitations, as pretrained LLMs frequently misconstrue human intentions or generate harmful informations \cite{41,42,43}. To address these limitations in our PatentGPT models, we utilized alignment techniques including Supervised Finetuning (SFT) and Reinforcement Learning from Human Feedback (RLHF). We collected over 43k SFT samples and trained pretrained PatentGPT models on them to construct the SFT models. Next, we applied the RLHF process to the SFT models to further improve their performance in the IP Domain. The conductions of SFT and RLHF processed on PatentGPT models are shown in Figure \ref{fig3}.

\subsection{Supervised Finetuning}
\textbf{Data:} In total, we used 30,000 general instructions ($\mathcal{D}_{gen}$) and 13,000 IP-specific instructions ($\mathcal{D}_{exp}$) for conducting SFT on the pretrained PatentGPT models. $\mathcal{D}_{gen}$ is composed of sampled data from several public datasets, including FLAN datasets \cite{44}, COIG datasets \cite{45}, Firefly \cite{46}, BELLE \cite{47}, MOSS \cite{48}, and Ultrachat \cite{49}. $\mathcal{D}_{exp}$ consists of IP-specific instructions annotated by patent examiners and other experts in technics. $\mathcal{D}_{exp}$ not only contains Q\&A data in the IP domain and technical domains, but also includes supervised data specially designed to support retrieval augmented generation (RAG), and data supporting several IP application scenarios, including technical means summaries, implementation summaries, and patent comparisons, among others. The final dataset for SFT is $\mathcal{D}_{SFT}$,\ where $\mathcal{D}_{SFT}=\ \mathcal{D}_{gen}\ {\cup\ \mathcal{D}}_{exp}$.\\ 
\\
\textbf{Training:} For SFT, we concatenated all the instructions and outputs from $\mathcal{D}_{SFT}$. A special token was utilized to separate the instructs and outputs. All data from $\mathcal{D}_{SFT}$ was used to conduct SFT on PatentGPT-0.5 and PatentGPT-1.0-MoE, but only data with sequence length shorter than 4096 tokens was used to conduct SFT on PatentGPT-1.0-Dense.\\
\\
Inspired by Wang et al. \cite{50}, we utilized a weighted autoregressive objective and zeroed out the loss on tokens from the user instructions to better align with human intentions. Then loss function can be expressed as:\\
\begin{equation}
\begin{split}
    &\mathcal{L}_{SFT}\left(\Theta\right)=\mathbb{E}_{x\sim\mathcal{D}_{SFT}}\left[-\alpha\sum_{i\in{output}}\log{p}\left(x_i|x_0,\ x_1,.\ .\ .\ ,x_{i-1};\Theta\right)\right]\\
    &\alpha=\begin{cases}
    1, \ if\   x\in{D_{exp}}\\
    0.1, \ if\  x \in{D_{gen}}\\
    \end{cases}
\end{split}
\end{equation}
where $\Theta$ represents the model's parameters, $\mathcal{D}_{SFT}$ is the fine-tuning dataset, $x=\left(x_0,\ x_1,.\ .\ .\ ,x_{i-1} \right)$ represents the tokenized input sequence, output dominates the tokens that belong to the output segments. \\ 
\\
We employed a cosine learning rate schedule with a pick learning rate of $2\times{10}^{-5}$, a warmup of 10 \% of the total steps, a batch size of 64 for PatentGPT-0.5 and PatentGPT-1.0-MoE, and a batch size of 128 for PatentGPT-1.0-Dense. Finally, we fine-tuned the models for 3 epochs in total.
\subsection{Reinforcement Learning from Human Feedback}
RLHF is a training procedure applied to a fine-tuned language model to further align model behaviors with human preference and instruction following. We applied RLHF to the PatentGPT models, following the approaches of Bai et al. \cite{51}.\\
\\
\textbf{Reward model (RM):} We collected 100k human preference data for reward modeling, including 20k expert annotations and 80k AI-generated annotations. Each annotation consists of a prompt followed by several responses generated by our PatentGPT models of different sizes, and commercial LLMs such as GPT-4 and ChatGPT-3.5, to enhance response diversity. For an expert annotation, an annotator ranked the responses from best to worst following standard annotation guidelines, and constructed comparison pairs based on this ranking. In the AI-generated annotation process, responses were ranked by GPT-4 in accordance with the RLAIF approach proposed by Lee et al. \cite{52}.\\
\\
According to Lee's results \cite{51}, larger RMs achieve better performance for reward modeling. Therefore, we adopted the pretrained model of PatentGPT-1.0-Dense to initialize the parameters of the RM. This RM will be used for reinforcement learning in all PatentGPT models. The end token in each sample was aligned through left-padding, and two MLP layers were added to RM, enabling the model to output a scalar score that indicates human preference. Finally, we used a binary ranking loss to optimize the RM, which is expressed by\cite{53}:
\begin{equation}
\mathcal{L}_{ranking}=-\log\left(\sigma\left(r_\theta\left(x,y_c\right)-r_\theta\left(x,y_r\right)\right)\right)
\end{equation}
where $r_\theta\left(x,y_c\right)$ represents the scalar score output for a given prompt $x$ and its corresponding annotator-preferred response $y_c$, $y_r$ denotes the response that is rejected, and $\sigma$ denotes the sigmoid function.
Reinforcement learning (RL): To stabilize RL training, we used proximal policy optimization (PPO) \cite{54} with a reward signal provided by the RM score \cite{9}. During the RLHF training process, both the actor model and the reference model were initialized with the SFT models described in Section 3.1, while the critic model was initialized with RM. In each step, the actor model was required to generate four responses from a given prompt, and the response with the highest reward signal, provided by the RM scores, was selected to optimize the actor model.\\
For all models, we used the AdamW optimizer with $\beta_{1} = 0.9$, $\beta_{2} = 0.95$, and $eps = 10^{-5}$. A batch size of 64 and a KL divergence coefficient of 0.01 were used to train the RLHF model. The generation parameter, top p, was set to 0.9. The learning rates of the actor and critic models followed a cosine learning rate schedule with a peak learning rate of $1\times{10^{-6}}$ and decaying to a final learning rate of $1\times{10}^{-7}$.
\section{Evaluation Results}
In order to evaluate the general capabilities of LLMs objectively and effectively, benchmarks such as Truthful QA \cite{55}, LongBench \cite{56}, and MMLU \cite{57} have been proposed in previous works. On the other hand, considering the diverse demands of various industries, domain-specific benchmarks such as CoderEval \cite{58}, PubMedQA \cite{59}, and FinBench \cite{60} also have been introduced as evaluation standards for domain-specific LLMs. Recently, Ni et al. \cite{61} introduced an IP-oriented benchmark called MoZIP, which utilizes IP-related open question answering and multiple-choice questions to evaluate models' performance in IP-oriented knowledge and their proficiency in matching patent texts. However, MoZIP does not involve tasks such as patent specification drafting, patent classification, and summarizing key technical information from patent specifications, which are very important for the application of LLMs in the IP domain.\\
\\
In this section, we first introduce PatentBench, a comprehensive benchmark focusing on the IP domain. Then, we present the evaluation results of the PatentGPT models, as well as those of ChatGPT-3.5-turbo and GPT-4 on PatentBench, MoZIP, MMLU, and C-Eval, respectively.
\subsection{PatentBench}
To compensate for the lack of specialization in the IP domain of open-source benchmarks, we proposed PatentBench which for the first time incorporates tasks related to patent agency and examination process such as patent specification drafting, patent classification, and summarizing key technical points of patents into a benchmark for evaluating IP-specific LLMs. We plan to open-source PatentBench in the fourth quarter of this year, contributing to the development of the open source communities and the IP industry. The main evaluation tasks of PatentBench are as follows:\\
\\
\textbf{Patent\_QA:} We constructed the Patent\_QA dataset to evaluate if an LLM understand IP-related concepts and regulations. The instructions were collected from product experts, top sellers of Patsnap and FAQs from the official websites of major national IP offices and the leading IP education websites. Each instance consists of a prompt with one or two questions and the corresponding reference answers as the output. The total number of questions is 250 for Chinese and 250 for English. \\
\\
\textbf{Patent\_Writing:} Patent\_Writing is a dataset designed to evaluate the ability of LLMs to draft patent texts. It involves drafting patent specifications based on technical disclosure documents, rewriting and refining various sections within the patent descriptions, as well as supplementing the claims. All instances in this dataset were manually constructed by annotators using tools assisted by LLMs based on patent specifications in either English or Chinese. The total number of instances in this dataset is 300.\\
\\
\textbf{Patent\_Classification:} According to the classification rules of the International Patent Classification (IPC), patents can be divided into 118 categories based on the Section and Class of the IPC number. We have uniformly selected 1,180 patents from the Chinese and United States patent databases across these 118 categories. The instructions for this dataset require the LLMs to generate the correct categories based on the abstract and claims of each patent.\\
\\
\textbf{Patent\_Summary:} This dataset comprises four sub-tasks: technical effects, technical problems, technical means, and patent abstracts. Each sub-task contains 300 instances. First, we trained an extraction model to accurately extract paragraphs involving technical effects, technical problems, and technical means. Then we inputted the extracted paragraphs along with carefully designed prompts into GPT-4, requesting GPT-4 to summarize the technical effects, technical problems, technical means, or the abstracts of the patents based on the input contents. Each instance in the dataset consists of a prompt accompanied by extracted paragraphs and a task instruction, as well as the manually refined response from GPT-4 as the output.\\
\\
\textbf{Patent\_Reasoning:} We have created this dataset to evaluate the capability of LLMs in reasoning. The tasks covered by this dataset include patent infringement judgment, reasoning sentences' logical relationships, and mathematical and logical reasoning tasks. All instances in this dataset were semi-manually constructed in both English and Chinese with the assistance of GPT-4. The total number of instances in this dataset is 247.\\
\\
\textbf{Patent\_Correction:} First, annotators were required to transcribe patent specifications. Next, we checked these transcriptions to identify any sentences containing spelling errors. We used this data to construct 300 test samples designed to challenge LLMs to correct misspelled characters or words. Each sentence contains only one incorrect Chinese character or English word. Responses of LLMs that exactly match the corrected versions provided by the annotators are regarded as correct.\\
\\
\textbf{Patent\_Translation:} We collected users' feedback on incorrect translations from Patsnap's products. Based on this data, we constructed the Patent\_Translation dataset to assess the performance of LLMs in translating patent texts. Each instance in the dataset consists of a prompt and an output, where the prompt is made up of a text in original language and an instruction asking LLMs to translate the text into the target language. We have completely constructed 1,000 instances, of which 500 instances require the LLMs to translate Chinese texts into English, and the other 500 instances require the reverse translation from English to Chinese.
\subsection{Results on PatentBench}
\begin{figure}[!h]
\centering
\includegraphics[width=0.55\textwidth]{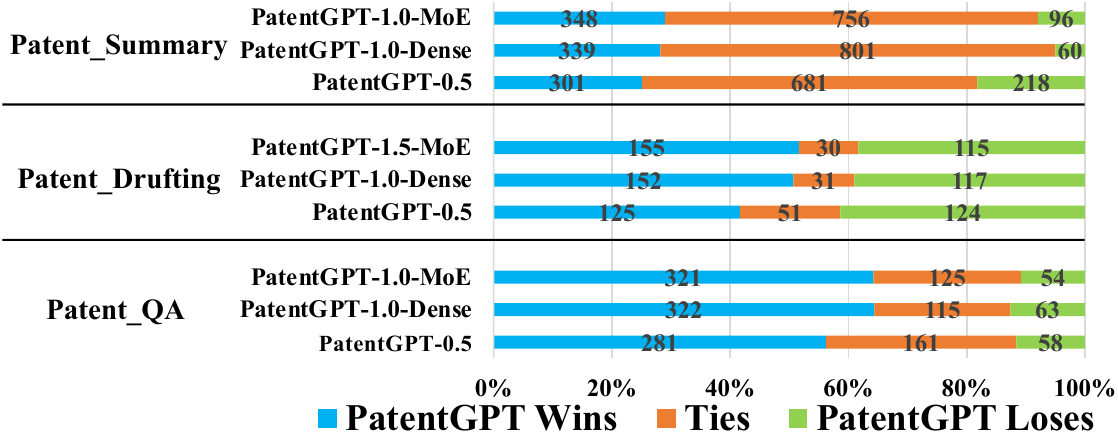}
\includegraphics[width=0.4\textwidth]{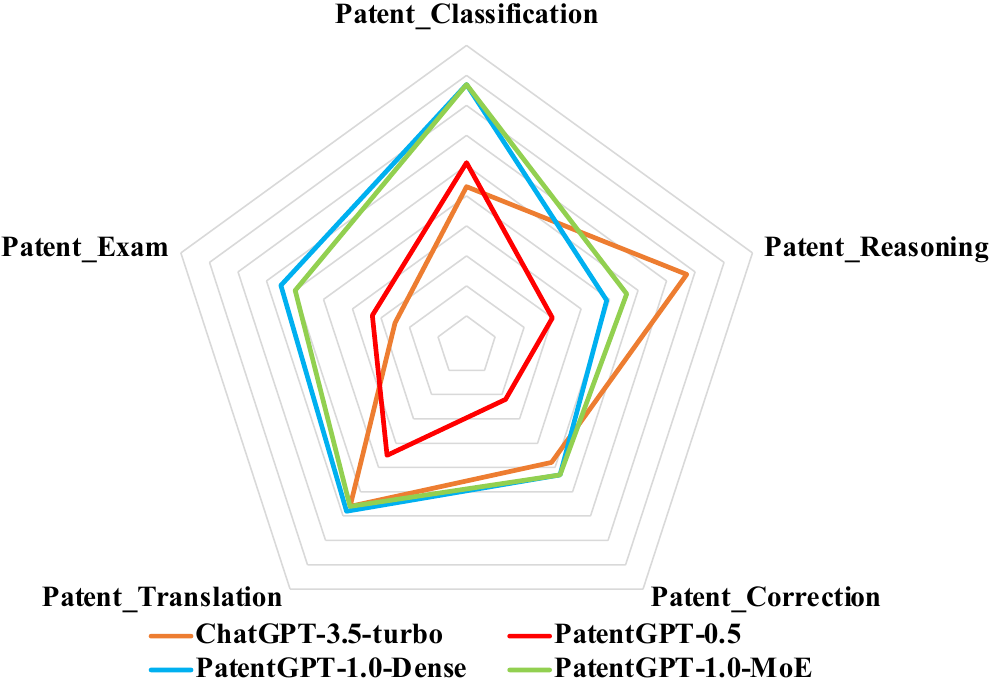}
\caption{\label{fig4}
Zero-shot performance of PatentGPT models on PatentBench: The left panel illustrates patent summarization, drafting and IP-oriented open question answering capabilities of PatentGPT models in comparison with GPT-3.5-turbo, as evaluated automatically by GPT-4. The right panel shows the classification, examination, translation, text correction, and reasoning abilities assessed based on metrics widely used in NLP.}
\end{figure}

We first evaluated zero-shot performance of our PatentGPT models on the PatentBench using both GPT-4 and the metrics widely used in NLP. To assess the models' summarization, writing, and conversational abilities, we employed GPT-4 as a judge by inputting a prompt that required it to compare the outputs from two assistants (GPT-3.5-turbo and one of our PatentGPT models) and to generate scores based on well-designed guidelines. Each pair of outputs from different models was evaluated twice with their positions in the prompt reversed, and the final scores were calculated by averaging, aiming to eliminate influences associated with position bias. Finally, we statisticed on the number of wins, loses, and ties across the two models, and the results is shown in Figure \ref{fig4}. Our PatentGPT models show significant improvements over ChatGPT-3.5-turbo in terms of drafting, and IP-oriented open question answering, indicating that our PatentGPT models have the potential to serve as patent assistants, thus helping humans in drafting patent specifications, reading patents, and understanding patent laws and regulations. In summarization tasks, the proportion of ties is very high, suggesting that the gap in summarization ability between ChatGPT-3.5-turbo and our PatentGPT models is difficult to be evaluated by GPT-4.\\
\\
We subsequently evaluated classification, examination, translation, text correction, and reasoning abilities based on different metrics. The classification capability of the models was measured by the F1 scores, the examination ability by accuracy, the translation quality by BLEU scores, and the reasoning and text correction abilities by EM scores \cite{62}. Besides PatentGPT-0.5, our PatentGPT models have outperformed ChatGPT-3.5-turbo in capabilities other than reasoning. These results demonstrate the effectiveness of our pre-training and the advanced nature of our PatentGPT models in the IP domain.\\
\\
To further demonstrate the expertise of the PatentGPT models in the IP domain, we tested PatentGPT models using the 2019 China Patent Agent Qualification Examination. For convenience in statistics, we simplified the score calculation rules for the patent agent exam. We determined that for the 200 multiple choice questions in patent law and related laws, each correct answer will be awarded 0.5 points, achieving a perfect score of 100 if all questions are answered correctly. Figure 5 depicts the exam scores of ChatGPT-3.5-turbo, GPT-4-1106-preview and the PatentGPT models on the patent agent exam. Results show that all commercial general LLMs failed to pass the cutoff of 60 points for the patent agent exam, whereas PatentGPT-1.0-Dense and PatentGPT-1.0-MoE scored 65 points and 60 points respectively, approaching the level of an IP expert. This result reveals the shortcomings of general LLMs in the IP domain, and once again emphasizes the effectiveness and necessity of pretraining domain specific LLMs.\\
\begin{figure}[!h]
\centering
\includegraphics[width=0.8\textwidth]{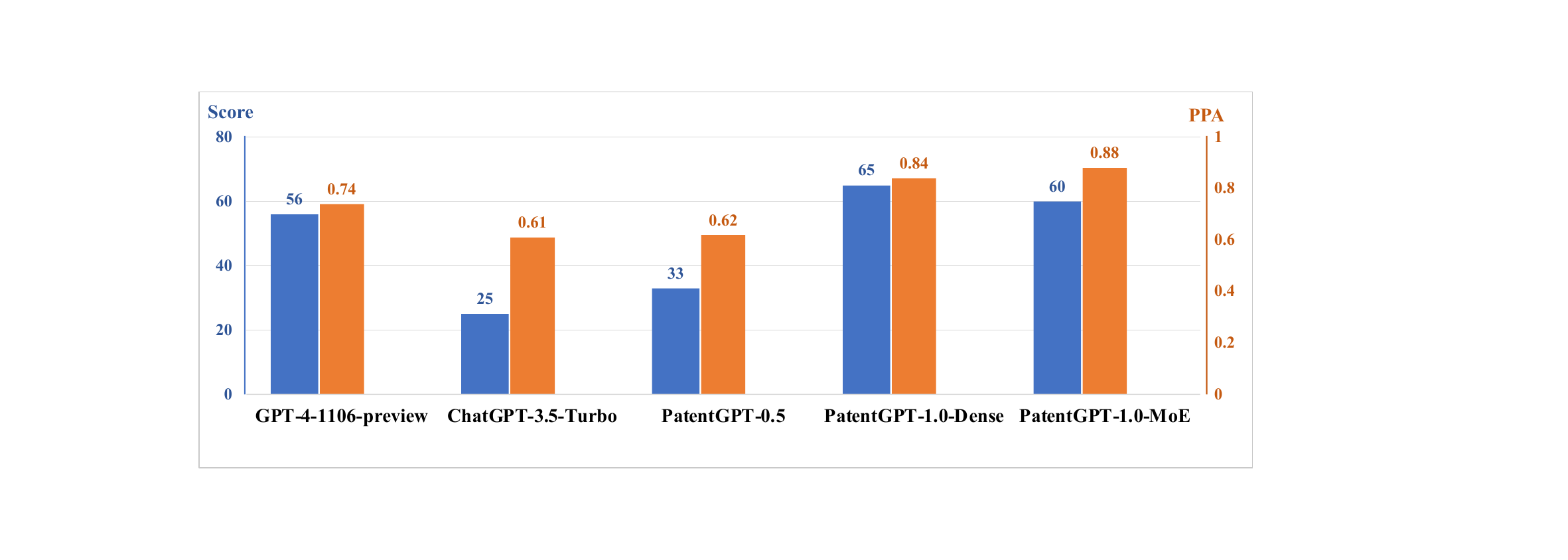}
\caption{\label{fig5}The performance of all models on the 2019 China Patent Agent Qualification Examination and their corresponding PPA scores.}
\end{figure}\\
\\
It should be particularly emphasized that patent agent exam results presented in Figure \ref{fig5} aren’t merely achieved by integrating IP knowledge into the models. We have also enhanced the models’ ability to eliminate the interference of option order through a series of data augmentation techniques, and this ability can be expressed in PPA, suggested by Robinson et al. \cite{63}. Through our data augmentation processes, the PPA of PatentGPT-1.0-Dense significantly outperforms that of GPT-4-1106-preview, implying a higher score ceiling for PatentGPT-1.0-Dense. 
\subsection{Results on MoZIP Benchmark}
\begin{figure}[!h]
\centering
\includegraphics[width=0.65\textwidth]{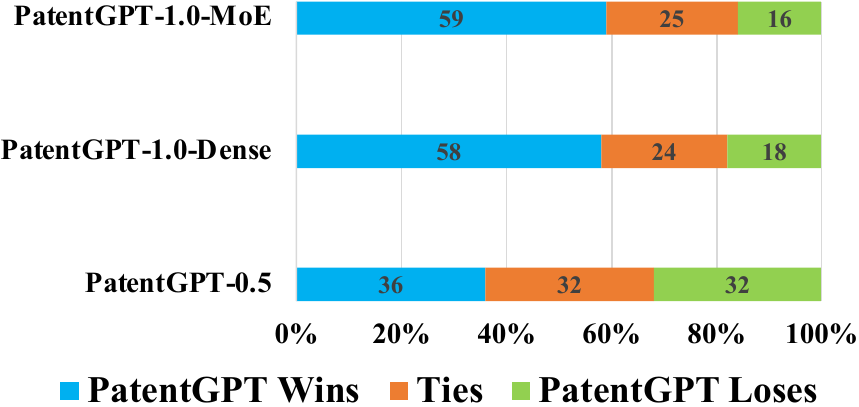}
\caption{\label{fig6}The number of questions on IPQA that PatentGPTs wins, ties or loses, compared to ChatGPT-3.5-turbo. }
\end{figure}

In addition to evaluating our PatentGPT models on the PatentBench, we also assessed them on MoZIP \cite{61}, which is a public benchmark specifically designed for the IP industry. MoZIP benchmark includes three challenging tasks: IP multiple-choice quiz (IPQuiz), IP question answering (IPQA), and patent matching (PatentMatch). GPT-3.5-turbo and GPT-4-1106-preview were used as baselines to evaluate the PatentGPT models. 
The 5-shot results on IPQuiz are shown in Table \ref{table2}. PatentGPT-1.0-Dense and PatentGPT-1.0-MoE outperform GPT-3.5-turbo, indicating that these PatentGPT models demonstrate a better understanding of the fundamental concepts and regulations of IP. The PatentGP-1.0 model achieved a score of 79.9, which is 2.1 points higher than the GPT-4-1106-preview, and once again performs the best among all the models, which is consistent with the results of the China Patent Agent Qualification Examination shown in Figure \ref{fig5}. This result suggests a significant advantage of PatentGPT-1.0-Dense over generic models in the IP domain.
We also notice that, in this experiment, the performance gaps between these models were not as significant as those revealed in Figure \ref{fig5}, which can be attributed to the fact that tasks in IPQuiz were clearly easier than those in the 2019 China Patent Agent Qualification Examination.\\
\begin{table}[!h]
\centering
\begin{tabular}{cccc}
\hline
Model & IPQuiz-EN&IPQuiz-ZH&Average\\
\hline
ChatGPT-3.5 turbo&68.1&66.0&67.1\\
GPT-4-1106-preview&74.9&80.6&77.8\\
PatentGPT-0.5&52.0&64.2&58.1\\
PatentGPT-1.0-Dense&\textbf{78.9}&\textbf{80.9}&\textbf{79.9}\\
PatentGPT-1.0-MoE&77.0&75.2&76.1\\
\hline
\end{tabular}
\caption{\label{table2}5-shot performance of all models on IPQuiz. IPQuiz-EN and IPQuiz-ZH comprise all test samples written in English and Chinese, respectively.}
\end{table}\\
Table \ref{table3} additionally illustrates that all PatentGPT models outperform GPT-3.5-turbo on PatentMatch, suggesting that compared to GPT-3.5-turbo, our models are superior at grasping the relationships between various patents. This capability is beneficial in assisting patent examiners in identifying potential infringement contents in a patent. It is worth noting that PatentGPT-0.5 has only 13 billion parameters, yet it has achieved a 15.6 point improvement on PatentMatch compared to GPT-3.5-turbo that has far more parameters. Furthermore, PatentGPT-1.0-Dense has an average score of 69.1, which is 2.8 points higher than that of GPT-4-1106-preview. These results indicate that our meticulously prepared pre-training data and strategies effectively enhance the model's performance in the IP domain.\\
\begin{table}[!h]
\centering
\begin{tabular}{cccc}
\hline
Model&PatentMatch-EN&PatentMatch-ZH&Average\\
\hline
ChatGPT-3.5 turbo&34.6&43.0&38.8\\
GPT-4-1106-preview&\textbf{72.3}&60.3&66.3\\
PatentGPT-0.5&49.5&59.2&54.4\\
PatentGPT-1.0-Dense&66.2&\textbf{72.0}&\textbf{69.1}\\
PatentGPT-1.0-MoE&70.6&60.8&65.7\\
\hline
\end{tabular}
\caption{\label{table3}5-shot performance of all models on IPQuiz. IPQuiz-EN and IPQuiz-ZH comprise all test samples written in English and Chinese, respectively.}
\end{table}\\
\\
During the IPQA evaluation, all questions in this dataset were initially answered by ChatGPT-3.5-turbo, GPT-4-1106-preview, and the PatentGPT models. Subsequently, annotators were asked to select which answer they considered superior, or to determine whether there was a significant difference between the two answers generated by the different models. If no distinction was apparent, then results were ties.\\
\\
Figure \ref{fig6} shows the number of questions on IPQA that PatentGPTs wins, ties or loses, compared to ChatGPT-3.5-turbo. All PatentGPT models achieve superior performance, despite having fewer parameters than ChatGPT-3.5-turbo. These results are consistent with those assessed by GPT-4 and presented in Figure 4. They once again confirm that human experts prefer the PatentGPT models over ChatGPT-3.5-turbo for addressing IP issues, demonstrating that PatentGPT models generate more professional answers with fewer hallucinations. \\
\subsection{Results on MMLU and C-Eval}
In the process of a patent specification being drafting or reviewed, common sense knowledge in areas such as physics, chemistry, and engineering are involved, along with patent laws. Therefore, we evaluated the scientific and technological common sense of our PatentGPT models using MMLU and C-Eval, which are widely used to evaluate performance of LLMs. We assessed the performance of the PatentGPT models and ChatGPT-3.5-turbo on the test set of MMLU and the validation set of C-Eval using zero-shot. The answering accuracy of each model is summarized in Table \ref{table4}, and the evaluation results of each subject are presented in Appendix \ref{B}. It demonstrates that PatentGPT-1.0-MoE performs comparably to GPT-3.5-turbo on both MMLU and C-Eval, suggesting that our PatentGPT models can not only resolve issues in patent laws but also excel in dealing with tasks such as patent drafting, patent comparison, patent classification, and technical word extension, which concern common sense understanding in various subjects.
\begin{table}[!h]
\centering
\begin{tabular}{cccc}
\hline
 Model&MMLU&C-Eval&Everage\\
\hline
PatentGPT-0.5&	51&	45&	48\\
PatentGPT-1.0-Dense&	61&	46&	53.5\\
PatentGPT-1.0-MoE&	59&	\textbf{52}&	55.5\\
ChatGPT-3.5-turbo&	\textbf{66}&	47&	\textbf{56.5}\\
\hline
\end{tabular}
\caption{\label{table4}Performance of PatentGPT models on MMLU and C-Eval compared to ChatGPT-3.5 turbo.}
\end{table}
\subsection{Results on inference resources consumption}
The evaluation results presented in Sections 4.1-4.3 demonstrate that PatentGPT-1.0-MoE slightly underperforms PatentGPT-1.0-Dense but achieves performance comparable to that of GPT-4. This indicates that both PatentGPT-1.0-Dense and PatentGPT-1.0-MoE have the potential to serve as alternatives to GPT-4 in the IP industry. In addition to performance, the response latency and operational costs of the models are critical for their commercial viability. Therefore, we performed the 4-bit quantization cross all PatentGPT models and employed Text-Generation-Inference(TGI)-1.4 to assess their resource consumption when they output the first token. All experiments were conducted on a server equipped with a NVIDIA A100 80GB GPU.
\begin{figure}[!h]
\centering
\includegraphics[width=0.95\textwidth]{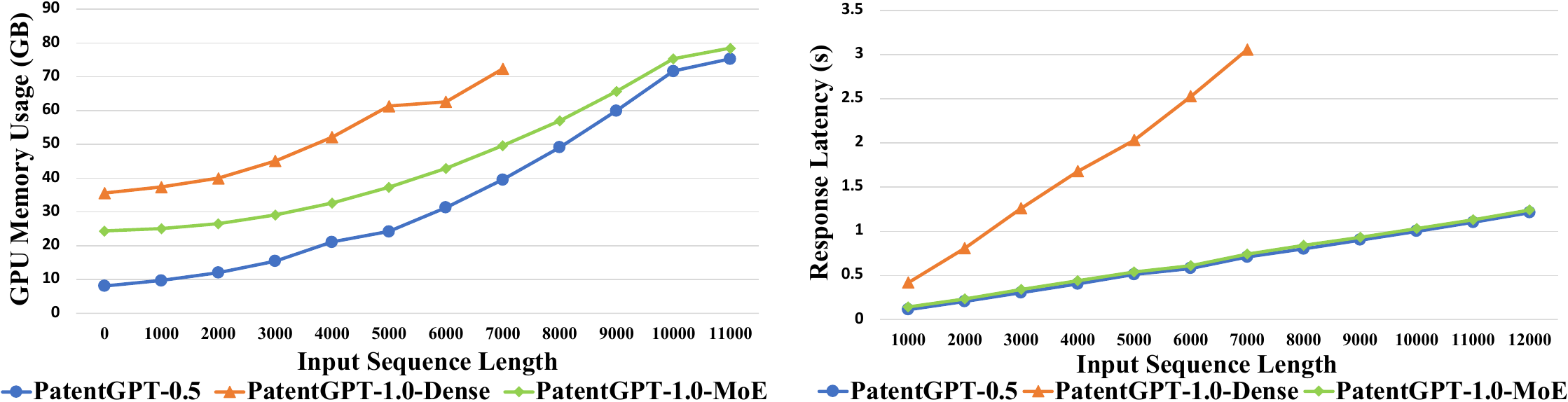}
\caption{\label{fig7}The correlation between input sequence length and resource consumption during the inference stage. The left panel illustrates the variation of GPU memory usage versus the input sequence length. The right panel shows the variation of response latency versus the input sequence length. }
\end{figure}\\
Figure \ref{fig7} illustrates the correlation between input sequence length and resource consumption during the inference stage. The results show that the GPU memory usage of all models is positively correlated with both their number of parameters and the length of the input sequence. Among them, PatentGPT-1.0-MoE exhibits the smallest increase in GPU memory usage as the input sequence length increases. This indicates that PatentGPT-1.0-MoE is more resource-efficient in applications that involve long contextual scenarios. Additionally, PatentGPT-0.5 and PatentGPT-1.0-MoE have the same response latency, which is significantly shorter than that of PatentGPT-1.0-Dense, suggesting that PatentGPT-0.5 and PatentGPT-1.0-MoE have relatively lower computation requirements. The smaller dimensions of the MLP layers and multi-head attention layers are responsible for these results.\\
\\
The experiment results also reveal that under conditions where model performance levels are nearly equivalent, PatentGPT-1.0-MoE tends to outperform in terms of inference efficiency. A particular point of concern for IP models is the computation resource management of long contexts sequence, which remains a significant challenge. The design of PatentGPT-1.0-MoE allows for handling long context and maintaining high performance levels without undue strain on system resources, offering a potentially superior compromise. This could mark a pivotal step in the deployment of advanced IP-oriented models in environments where resource management is just as critical as the capacity to handle complex and lengthy data sequences. These findings underscore the viability of SMoE models like PatentGPT-1.0-MoE in the IP industry and the importance of selecting the model architecture to develop powerful, efficient, and professional-use solutions.
\section{Conclusions and Future Works}
In this report, we introduced methodologies for training IP-oriented LLMs and establish the most comprehensive benchmark in the IP domain, the PatentBench. The main conclusions of this report are summarized as follows:\\
\\
First, we proposed a standard training procedure for LLMs in the IP domain, including data preprocessing, pretraining, alignment, and evaluation. Under the guidance of this procedure, we trained the PatentGPT models. The results of the experiments indicate that our PatentGPT models demonstrated overall performance that surpasses that of ChatGPT-3.5-turbo in the IP domain. Moreover, PatentGPT-1.0-Dense, with 70 billion parameters, even exhibits superior performance compared to GPT-4 in this field, indicating the effectiveness of the methodology we employed.\\
\\
Secondly, we compared the performance of PatentGPT models that were pre-trained using open-source base models with various parameter scales and across different frameworks. After a comprehensive evaluation of these models’ performance on IP-oriented tasks and their computational resource consumption, we found that IP-oriented LLMs with 70 billion parameters tended to perform better on tasks in the IP domain, while the SMoE model with 47 billion parameters exhibited a better cost-performance ratio on long-text tasks, achieving a better tradeoff between model quality and inference efficiency.\\
\\
Finally, we provided a benchmark closer to the use cases of LLMs in the IP domain, called PatentBench, which serves a reference for comprehensive evaluation of LLMs in the IP domain.\\
\\
Future work will focus on enhancing long-context support, aiming to achieve the ability to support up to 128k tokens to meet more diverse IP-oriented scenarios. Additionally, we will prioritize the accumulation of English pretraining corpora and SFT data to further improve our models’ performance in English.
\printbibliography
\appendix
\newpage
\section{Examples of Synthetic Data}
\textbf{Combination of Patent Text and File Wrapper:}\\
In this section, we introduce the method for constructing synthetic data using structured data from various sources along with the corresponding attribute labels. \\
Taking the generation of synthetic data with File Wrappers and patents as an example:\\
We first extracted the patent numbers of the Patent under Examination and the Public Patent used as comparisons from the corresponding File Wrapper using named entity recognition (NER) to construct attribute labels. \\
Then, utilizing these attribute labels, we located the claims of the Patent under Examination and the claims and description of the Public Patent from the patent database. These elements were concatenated with the sections of the File Wrapper that expound the examination comments, forming synthetic data for pretraining. Table \ref{table5} and Figure \ref{table6} illuminate the English and Chinese examples of the mentioned process, respectively.\\
\\
\textbf{Extraction of Similar Corpus Pairs in the File Wrapper:}\\
In addition to combining patent texts and File Wrapper into a training corpus, we can also extract X-file text pairs (text pairs extracted from the patent under examnation and the public patent, respectively, representing the existence of technological connections between the two patents) from the File Wrapper, and the X-file text pairs can also be combined with other patent texts to form training data or instruction data. Table 7 shows the logic of generating English X-file text pairs, Table 8 shows the logic of generating Chinese X-file text pairs, and Table 9 shows the samples of X-file text combined with other patent text to form training data.\\
Firstly, we extracted the description text for patent comparison by the examiner from the File Wrapper; secondly, we extracted the original text (at the level of large paragraphs) for patent comparison mentioned in the File Wrapper from the patent under examnation and the public patent (the patent used for comparison); finally, we extracted the original sentence from the two patents according to the description text for patent comparison mentioned in the File Wrapper and the original text (at the level of large paragraphs) of the patent.\\
With training the similar corpus pairs (X-file pairs) that showed in Table 9\label{table9}, PatentGPT can be realized to capture the sentence pairs with the closest technical expression in patent comparisons, which meanwhile that the later application of the two patents may be at risk of infringement. This capability can greatly improve the efficiency and accuracy of IP practitioners in designing the scope of patent protection.
\label{A}
\newpage
\begin{table}[!h]
\centering
\begin{tabular}{|p{15cm}|}
\hline
\fcolorbox{cyan!40}{cyan!40}{\parbox{0.985\linewidth}{\textbf{Patent under Examnation(PN:US 2023/0333692 A1):}\\
\textbf{Claims:}\\
1. An electronic device comprising: a display layer; a sensor layer on the display layer, and comprising: a plurality of electrodes comprising a first electrode, and a second electrode adjacent to the first electrode; …  in the second electrode by the input device.\\
…\\
20. The electronic device of claim 16, wherein the plurality of electrodes further comprise: a third electrode interposed between the first electrode and the second electrode, and a fourth electrode interposed between the first electrode and the third electrode...}}
\fcolorbox{green!40}{green!40}{\parbox{0.985\linewidth}{\textbf{Public Patent(PN: US 2014/0078104 A1, Authors: De Foras et al.)}\\
\textbf{Claims:}\\
1. A display device, comprising: a display panel comprising a first display substrate and a second display substrate facing the first display substrate … and the second sensing signals.\\
…\\
90. The display device of claim 89, wherein the display period is longer than the non-display period, and the first scan signal output part is configured to provide the first scan signals more than once to the first touch coils during the display period.\\
\\
\textbf{Description:}\\
CROSS REFERENCE TO RELATED APPLICATION\\
$\left[0001\right]$ This U.S. non-provisional patent application claims priority from and the benefit of U.S. Provisional Patent Application No. 61/701,100, filed on Sep. 14, 2012, Korean Patent Application No. 10-2013-0021423, filed on Feb. 27, 2013, Korean Patent Application No. 10-2013-0021426, filed on Feb. 27, 2013, and Korean Patent Application No. 10-2013-0055845, filed on May 16, 2013, which are hereby incorporated by reference for all purposes as if fully set forth herein. \\
…\\
$\left[0475\right]$ Although the exemplary embodiments of the present disclosure have been described, it will be apparent to those skilled in the art that various modifications and variations can be made in the present disclosure without departing from the spirit or scope of the disclosed subject matter. Thus, it is intended that the present disclosure cover the modifications and variations of the disclosed subject matter provided they come within the scope of the appended claims and their equivalents.}}
\fcolorbox{yellow!40}{yellow!40}{\parbox{0.985\linewidth}{\textbf{File Wrapper:}\\
DETAILED ACTION\\
Notice of Pre-AIA or AIA Status\\
…\\
Instant Claim 2: The electronic device of claim 1, wherein the driving circuit comprises an analog front end comprising: an amplifier configured to amplify the first input signal and the second input signal; and a filter configured to remove noise from a signal output from the amplifier. (“The first signal processor 530 (fig 28B) includes an amplifier, a noise filter, and an analog-to-digital converter. The amplifier amplifies the first sensing signals SS1(1) to SS1(r). The noise filter removes noises from the amplified first sensing signals SS1(1) to SS1(r).” (Lee, paragraph 296)) \\
…\\
Examier, Art Unit }}\\
\hline
\end{tabular}
\caption{\label{table5}An example of synthetic data in English: \\
texts highlighted with a blue background indicate claims from the Patent under Examination; texts highlighted with a green background represent the claims and description of the Public Patent; and texts highlighted with a yellow background denote examination comments from the File Wrapper.}
\end{table}
\newpage
\begin{table}[!h]
\centering
\begin{tabular}{|p{15cm}|}
\hline
\begin{CJK}{UTF8}{gkai}
\fcolorbox{cyan!40}{cyan!40}{\parbox{0.985\linewidth}{\textbf{Patent under Examnation(PN:CN111865430A):}\\
\textbf{Claims:}\\
1.一种光电接收器，其特征在于，所述光电接收器包括壳体、跨阻放大器、固定支架和光电转换元件，所述壳体设有安装面，所述跨阻放大器固定于所述安装面上，所述固定支架 ... 位于所述跨阻放大器在所述安装面的正投影内。\\
...\\
10.如权利要求1～6任意一项所述的光电接收器，其特征在于，所述承载部设有定位孔，所述光电转换元件嵌入所述定位孔中，以对所述光电转换元件进行定位。}}
\fcolorbox{green!40}{green!40}{\parbox{0.985\linewidth}{\textbf{Public Patent(PN:CN207366793U)}\\
\textbf{Claims:}\\
1.一种光接收器件结构，其特征在于，结构中包括芯片底座(31)、探测芯片(32)、放大器芯片(33)、热沉(34)和封帽(35)，具体的：...之间通过金丝焊接连接。\\
...\\
8.所述权利要求1所述的光接收器件结构，其特征在于，所述芯片底座(31)、热沉(34)和封帽(35)按照TO56的规格制作。\\
\\
\textbf{Description:}\\
技术领域\\
$\left[0001\right]$本实用新型涉及光器件技术领域，尤其涉及一种光接收器件结构。\\
...\\
$\left[0036\right]$本实用新型所提出的光接收器件的封装结构适用于背景技术中所提到的TO56，具体的：所述芯片底座31、热沉34和封帽35按照TO56的规格制作。\\
$\left[0037\right]$以上所述仅为本实用新型的较佳实施例而已，并不用以限制本实用新型，凡在本实用新型的精神和原则之内，所作的任何修改、等同替换、改进等，均应包含在本实用新型的保护范围之内。}}
\fcolorbox{yellow!40}{yellow!40}{\parbox{0.985\linewidth}{\textbf{File Wrapper:}\\
第一次审查意见通知书\\
本申请涉及一种光电接受器。经审查，现提出如下审查意见。 \\
...\\
1.权利要求1请求保护一种光电接收器，对比文件1（CN207366793U）公开了一种光接收器结构，并具体公开了以下技术特征（参见说明书5-11段，图3-5）：
本实用新型提供一种光接收器件结构（相当于“光电接收器”），结构中包括芯片底座31、探测芯片32（相当于“光电转换元件”）、放大器芯片33、热沉34（相当于“固定支架”）和封帽35（底座31和封帽35相当于“壳体”），具体的：...\\
...\\
基于上述理由，本申请的独立权利要求以及从属权利要求均不具备创造性。同时，说明书中也没有可以被授予专利权的实质性内容，因而本申请不具备被授予专利权的前景。...\\
...}}
\end{CJK}\\
\hline
\end{tabular}
\caption{\label{table6}An example of synthetic data in Chinese: \\
texts highlighted with a blue background indicate claims from the Patent under Examination; texts highlighted with a green background represent the claims and description of the Public Patent; and texts highlighted with a yellow background denote examination comments from the File Wrapper.}
\end{table}
\newpage
\begin{table}[!h]
\centering
\begin{tabular}{|p{15cm}|}
\hline
\begin{CJK}{UTF8}{gkai}
\fcolorbox{cyan!40}{cyan!40}{\parbox{0.985\linewidth}{\textbf{Related Descriptions from File Wrapper:}\\
...\\
Claims 1-5 and 7 are rejected under 35 U.S.C. 102(a)(1) as being anticipated by Lee (US 2014/0078104 A1).\\
\\
Instant Claim 1: An electronic device comprising: a display layer; a sensor layer on the display layer, (“The display device includes a display panel LDP (fig 1), a signal controller 100, a gate driver 200, a data driver 300, and a touch panel.” (Lee, paragraph 104) The display device, display panel LDP, and touch panel of Lee correspond to the electronic device, display layer, and sensor layer of the claim, respectively.) and comprising: a plurality of electrodes comprising a first electrode, and a second electrode adjacent to the first electrode;\\
...}}
\fcolorbox{green!40}{green!40}{\parbox{0.985\linewidth}{\textbf{Paragraph-level text pairs:}\\
\textbf{Claim 1 from Patent under Examnation(PN:US 2023/0333692 A1):}\\
1. An electronic device comprising: a display layer; a sensor layer on the display layer, and comprising: a plurality of electrodes comprising a first electrode, and a second electrode adjacent to the first electrode; and a plurality of crossing electrodes; and a driving circuit electrically connected to the sensor layer, and configured to control an operation of the sensor layer, wherein the sensor layer is configured to operate in a first touch mode to sense a first input signal based on a capacitance change, and a second touch mode to sense a second input signal of an input device configured to discharge a magnetic field, and wherein the driving circuit is configured to sense coordinates of the input device based on a signal profile, the signal profile being measured based on a first current having a first current direction formed in the first electrode by the input device, and a second current having a second current direction opposite to the first current direction formed in the second electrode by the input device.\\
\textbf{Description para 104 texts from Public Patent(PN:US 2014/0078104 A1 , author: Lee):}\\
$\left[0104\right]$The display device includes a display panel LDP, a signal controller 100, a gate driver 200, a data driver 300, and a touch panel. The touch panel includes a plurality of scan lines TL1 to TLi (“i” is any whole number greater than 1), a plurality of source lines RL1 to RLj (“j” is any whole number greater than 1), a first driver 400, a second driver 500, and a touch sensor 600. The signal controller 100, the gate driver 200, and the data driver 300 control the display panel LDP to generate an image. The first driver 400 and the second driver 500 control the touch panel, and the touch sensor 600 calculates coordinate information of input positions.
}}
\fcolorbox{yellow!40}{yellow!40}{\parbox{0.985\linewidth}{\textbf{Sentence-level text pairs:}\\
\textbf{Sentence from Patent under Examnation(PN:US 2023/0333692 A1):}\\
An electronic device comprising: a display layer; a sensor layer on the display layer\\
\\
\textbf{Sentence from Public Patent(PN:PN:US 2014/0078104 A1):}\\
The display device includes a display panel LDP, a signal controller 100, a gate driver 200, a data driver 300, and a touch panel.}}
\end{CJK}\\
\hline
\end{tabular}
\caption{\label{table7}An example of a English X-file text pair generation process: \\
texts highlighted with a blue background indicate the paragraphs in the file wrapper that are relevant to the patent comparison; texts highlighted with a green background represent the text pairs of the technically closest passages extracted from the patent under examnation and the public patent, respectively; and texts highlighted with a yellow background denote sentence-level text pairs extracted from the above text pairs of the technically closest passages.\\
X-file text pairs are parsed into multiple levels, e.g. paragraph-leve text pairs, sentence-level text pairs, etc. These text pairs can be used as training corpus for the model with additional information.}
\end{table}
\newpage
\begin{table}[!h]
\centering
\begin{tabular}{|p{15cm}|}
\hline
\begin{CJK}{UTF8}{gkai}
\fcolorbox{cyan!40}{cyan!40}{\parbox{0.985\linewidth}{\textbf{Related Descriptions from File Wrapper:}\\
...\\
1.权利要求1请求保护一种光电接收器，对比文件1（CN207366793U）公开了一种光接收器结构，并具体公开了以下技术特征（参见说明书5-11段，图3-5）：\\
本实用新型提供一种光接收器件结构（相当于“光电接收器”），结构中包括芯片底座31、探测芯片32（相当于“光电转换元件”）、放大器芯片33、热沉34（相当于“固定支架”）和封帽35（底座31和封帽35相当于“壳体”），具体的：\\
优选的，放大器芯片33具体为跨阻放大器（相当于“跨阻放大器”）。\\
所述热沉34位于所述底座上（相当于“固定支架固定连接壳体”），热沉34靠近芯片底座31中心的一侧固定所述探测芯片32（公开了“固定支架承载部，所述光电转换元件固定于承载部”），其中，所述探测芯片32的收光面与封帽35上的窗口在一直线上；\\
...\\
从图5中可以看出，探测芯片32位于放大器芯片33的上方（相当于“光电转换元件在所述安装面上的正投影位于所述跨阻放大器在所述安装面的正投影内”）。\\
...}}
\fcolorbox{green!40}{green!40}{\parbox{0.985\linewidth}{\textbf{Paragraph-level text pairs:}\\
\textbf{Claim 1 from Patent under Examnation(PN:CN111865430A):}\\
1.一种光电接收器，其特征在于，所述光电接收器包括壳体、跨阻放大器、固定支架和光电转换元件，所述壳体设有安装面，所述跨阻放大器固定于所述安装面上，所述固定支架固定连接所述壳体，所述固定支架设有背离所述跨阻放大器的承载部，所述光电转换元件固定于所述承载部，以使所述光电转换元件在所述安装面上的正投影位于所述跨阻放大器在所述安装面的正投影内。\\
\\
\textbf{Description para 5-11 texts from Public Patent(PN:CN207366793U):}\\
$\left[0005\right]$本实用新型提供一种光接收器件结构，结构中包括芯片底座31、探测芯片32、放大器芯片33、热沉34和封帽35，具体的：\\
$\left[0006\right]$所述热沉34位于所述底座上，热沉34靠近芯片底座31中心的一侧固定所述探测芯片32，其中，所述探测芯片32的收光面与封帽35上的窗口在一直线上；\\
$\left[0007\right]$所述放大器芯片33固定在所述底座上，所述放大器芯片33、芯片底座31上的管脚以及探测芯片32之间通过金丝焊接连接。\\
$\left[0008\right]$优选的，所述探测芯片32的收光面与封帽35上的窗口在一直线上，具体为：\\
$\left[0009\right]$所述封帽35中心的窗口上设置有透镜351，所述透镜351的焦点与所述探测芯片32的收光点重合。\\
$\left[0010\right]$优选的，所述探测芯片32固定在所述热沉34上，且所述放大器芯片33固定在所述芯片底座31上后，探测芯片32的管脚与所述放大器芯片33的管脚相距预设距离；其中，所述预设距离为500um～1000um。\\
$\left[0011\right]$优选的，放大器芯片33具体为跨阻放大器。}}
\fcolorbox{yellow!40}{yellow!40}{\parbox{0.985\linewidth}{\textbf{Sentence-level text pairs:}\\
\textbf{Sentence from Patent under Examnation(PN:CN111865430A):}\\
所述固定支架设有背离所述跨阻放大器的承载部，所述光电转换元件固定于所述承载部\\
\\
\textbf{Sentence from Public Patent(PN:CN207366793U):}\\
热沉34靠近芯片底座31中心的一侧固定所述探测芯片32}}
\end{CJK}\\
\hline
\end{tabular}
\caption{\label{table8}An example of a Chinese X-file text pair generation process: \\
texts highlighted with a blue background indicate the paragraphs in the file wrapper that are relevant to the patent comparison; texts highlighted with a green background represent the text pairs of the technically closest passages extracted from the patent under examnation and the public patent, respectively; and texts highlighted with a yellow background denote sentence-level text pairs extracted from the above text pairs of the technically closest passages. \\
X-file text pairs are parsed into multiple levels, e.g. paragraph-leve text pairs, sentence-level text pairs, etc. These text pairs can be used as training corpus for the model with additional information.}
\end{table}
\newpage
\begin{table}[!h]
\centering
\begin{tabular}{|p{15cm}|}
\hline
\begin{CJK}{UTF8}{gkai}
\fcolorbox{cyan!40}{cyan!40}{\parbox{0.985\linewidth}{\textbf{Sentence-level text pairs:}\\
\textbf{Sentence from Patent under Examnation(PN:US 2023/0333692 A1):}
An electronic device comprising: a display layer; a sensor layer on the display layer.\\
\textbf{Sentence from Public Patent(PN:PN:US 2014/0078104 A1):}
The display device includes a display panel LDP, a signal controller 100, a gate driver 200, a
data driver 300, and a touch panel.}}
\fcolorbox{green!40}{green!40}{\parbox{0.985\linewidth}{\textbf{Sample of Synthetic Training Data:}\\
\textbf{human:}\\
Here are two patents, compare these two patents and extract the sentence pairs from each which is technically closest in presentation.\\
Here is the claims and description from Patent US 2023/0333692 A1:\\
- - (Insert the text of the claims and specification of the patent US 2023/0333692 A1) - -\\
Here is the claims and description from Patent US 2014/0078104 A1:\\
- - (Insert the text of the claims and specification of the patent US 2014/0078104 A1) - -\\
\textbf{gpt:}\\
Here are the technically closest sentence pairs in these two patents：\\
1.\\
An electronic device comprising: a display layer; a sensor layer on the display layer（from US 2023/0333692 A1），\\
The display device includes a display panel LDP, a signal controller 100, a gate driver 200, a
data driver 300, and a touch panel （from US 2014/0078104 A1）；\\
2.\\
- - (Insert the other sentence pairs from these two patents) - -\\
...}}
\end{CJK}\\
\hline
\end{tabular}
\caption{\label{table9}An example of synthetic training data in English consisting of X-file sentence pairs (extract through File Wrapper and original patent text) and original patent text:\\
texts highlighted with a blue background represent the sentence-text pairs in the file wrapper that are relevant to the patent comparison; \\
and texts highlighted with a green background represent the training corpus consisting of sentence-text pairs, original texts of relevant patents, and designed contexts.}
\end{table}
\newpage
\section{Performance of LLMs on C-Eval and MMLU}
\label{B}
\begin{figure}[!ht]
\centering
\includegraphics[width=0.95\textwidth]{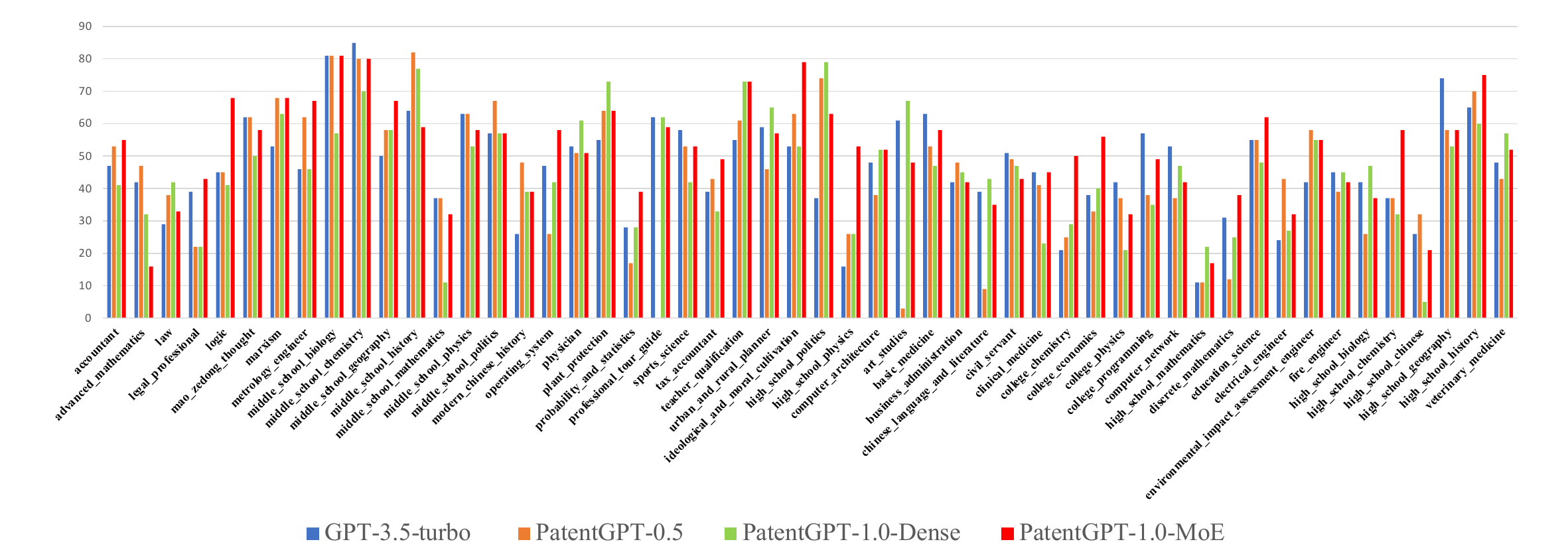}
\caption{\label{fig8}Performance of PatentGPT models and ChatGPT-3.5-turbo on C-Eval.}
\end{figure}
\begin{figure}[!ht]
\centering
\includegraphics[width=0.95\textwidth]{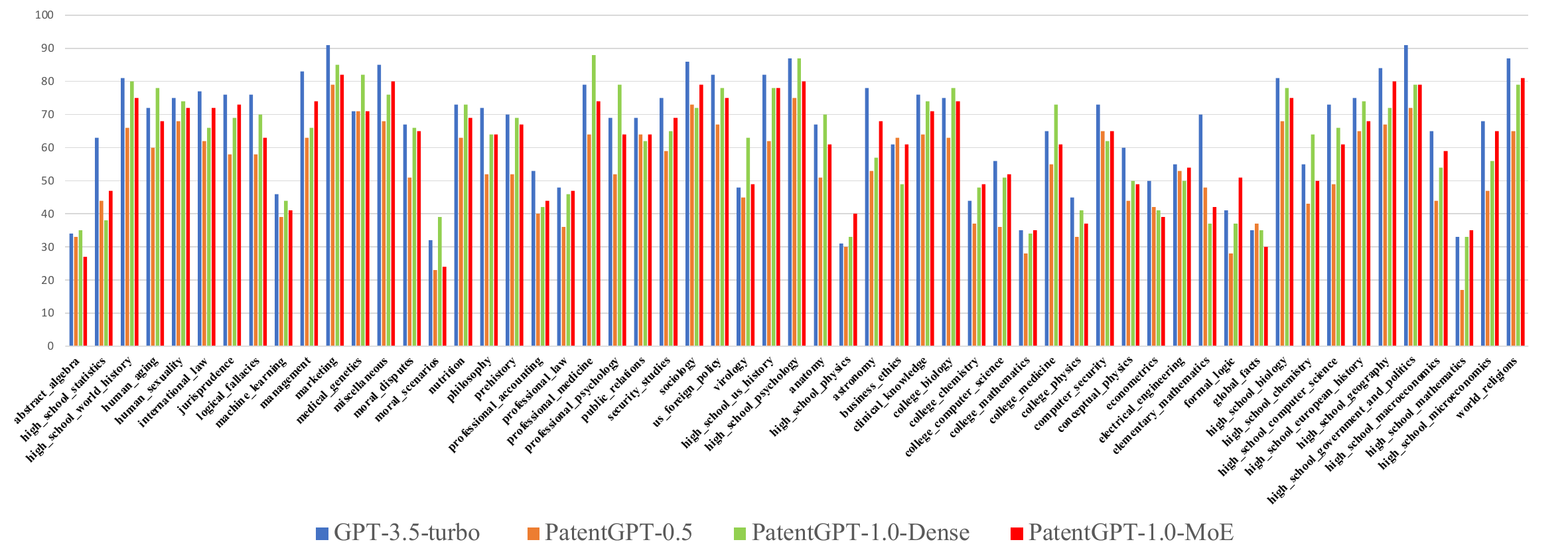}
\caption{\label{fig9}Performance of PatentGPT models and ChatGPT-3.5-turbo on MMLU.}
\end{figure}

\end{document}